\documentclass[letterpaper, 10 pt, conference]{ieeeconf}  

\IEEEoverridecommandlockouts                              

\usepackage{epsfig} 

\usepackage{mathtools} 
\usepackage{booktabs} 
\usepackage{tikz} 
\usepackage{times}
\usepackage{mathtools}
\usepackage{mathabx}
\usepackage{bm}
\usepackage{soul}
\usepackage{url}
\usepackage[hidelinks]{hyperref}
\usepackage[utf8]{inputenc}
\usepackage{graphicx}
\usepackage{amsmath}
\usepackage{booktabs}
\usepackage{algorithm}
\usepackage{algorithmic}
\usepackage[switch]{lineno}
\usepackage{xcolor}
\usepackage{graphicx} 
\usepackage{multirow} 
\usepackage{array} 
\usepackage{booktabs} 
\usepackage{diagbox}
\usepackage{graphicx}
\usepackage{rotating}
\usepackage{svg}
\usepackage{microtype}
\usepackage{subcaption}
\graphicspath{{FIGURES/}}
\svgpath{{FIGURES/}}

\title{\LARGE \bf

Integrating Perceptions: A Human-Centered Physical Safety Model for Human-Robot Interaction
}

\author{Pranav Pandey \and Ramviyas Parasuraman \and Prashant Doshi
\thanks{The authors are with the School of Computing, University of Georgia, Athens, GA 30602, USA. 
Author emails: {\small \{pranav.pandey,ramviyas,pdoshi\}@uga.edu}}
 }

\begin{document}

\maketitle

\begin{abstract}

Ensuring safety in human-robot interaction (HRI) is essential to foster user trust and enable the broader adoption of robotic systems. Traditional safety models primarily rely on sensor-based measures, such as relative distance and velocity, to assess physical safety. However, these models often fail to capture subjective safety perceptions, which are shaped by individual traits and contextual factors. In this paper, we introduce and analyze a parameterized general safety model that bridges the gap between physical and perceived safety by incorporating a personalization parameter, $\rho$, into the safety measurement framework to account for individual differences in safety perception.
Through a series of hypothesis-driven human-subject studies in a simulated rescue scenario, we investigate how emotional state, trust, and robot behavior influence perceived safety. Our results show that $\rho$ effectively captures meaningful individual differences, driven by affective responses, trust in task consistency, and clustering into distinct user types. Specifically, our findings confirm that predictable and consistent robot behavior as well as the elicitation of positive emotional states, significantly enhance perceived safety. Moreover, responses cluster into a small number of user types, supporting adaptive personalization based on shared safety models. Notably, participant role significantly shapes safety perception, and repeated exposure reduces perceived safety for participants in the casualty role, emphasizing the impact of physical interaction and experiential change.
These findings highlight the importance of adaptive, human-centered safety models that integrate both psychological and behavioral dimensions, offering a pathway toward more trustworthy and effective HRI in safety-critical domains.

\end{abstract}

\section{Introduction}

    As robots become more common in human environments, ensuring safety in physical human-robot interaction (pHRI) is critical. Although some robots are built to meet strict safety standards—limiting even worst-case collisions to minor injuries~\cite{haddadin2008collision}—these measures do not always ensure that people \textit{feel} safe. A robot can be physically safe yet still perceived as threatening or uncomfortable~\cite{nertinger2023influence}, highlighting the importance of \textbf{perceived physical safety}~\cite{kirschner2023influence}. If perceived safety is lacking, it can induce stress and hinder workplace acceptance~\cite{nertinger2023influence}. Despite this recognition, most HRI safety approaches remain focused on physical risk mitigation, with limited integration of psychological factors. Recent works have attempted to incorporate perceived safety by reducing robot speed, maintaining greater separation distances, or employing communicative cues~\cite{rossi2017user}, but these strategies are often applied as secondary modifications rather than part of a unified safety framework. This ad-hoc approach introduces trade-offs, as individual parameters like speed and distance interact in complex ways. Addressing the gap between ensuring \textit{actual} and \textit{perceived} physical safety remains a critical challenge, necessitating frameworks that seamlessly integrate both aspects for effective HRI.

    In this paper, our objective is to understand this gap by exploring the relationship between how humans psychologically assess risk and comfort with respect to robot dynamics, personal expectations, and objective safety metrics, and mitigate the gap between absolute physical safety and perceived safety by analyzing a unified \textbf{human-focused safety model}. The proposed model extends conventional physical safety constraints with an additional criterion derived from human safety perception. Our findings aim to inform the design and control of robots that are both physically safe and perceived as safe by their users. This study not only improves the design of safety frameworks but also enables the deployment of robotic systems in critical applications such as healthcare, disaster response, and public assistance, where user perception is paramount. 
    
    The key contributions of this paper are the following: $(1)$ we introduce a novel approach leveraging a proxemics-guided parameterized safety model to bridge classical physical safety metrics with human-aware perceived safety considerations, $(2)$ we empirically quantify the model’s personalization parameter through controlled user studies, demonstrating how human factors (such as emotional state and situational role) significantly affect perceived safety, and $(3)$ we present insights showing how the safety parameter exhibits distinct and general trends (for instance, differing between bystander and casualty roles). Finally, we discuss how incorporating these human-centered findings can inform perception-aware robot planning and control. Together, these contributions advance a more holistic approach to robot safety, prevent harm, and proactively foster human comfort and trust in HRI. A video demonstrating the experimental setup is available at \url{https://tinyurl.com/roman2025-video}.

\section{Background}
    
    Understanding safety in HRI requires differentiating between two key aspects: (1)~\textbf{Absolute Physical Safety}, defined by measurable parameters such as distance, velocity, and orientation of humans with respect to robots; and (2)~\textbf{Perceived Physical Safety}, the human’s psychological assessment of safety around robots based on subjective factors such as predictability and emotional responses. Although absolute safety models ensure compliance with regulatory standards, they often fail to capture the human-centric factors that affect trust and adoption. This section provides a foundation for these concepts, highlighting limitations of existing safety models and the need for integrating perceptions.

    \subsection{Absolute Physical Safety}

    Absolute physical safety (APS) prevents harm in HRI through objective metrics such as relative human-robot distance, velocity, and orientation, ensuring compliance with safety standards. Technological advances have enhanced safety through vision-based systems that utilize deep learning for collision detection~\cite{rodrigues2022modeling} and human tracking methods that integrate zone-based safety measures based on movement speed and robot response time~\cite{maria2022vision}. Advanced sensing technologies, such as RGB-D cameras, further improve detection capabilities by incorporating depth data into safety assessments. Computational models for risk assessment leverage keypoint-based monitoring to implement speed and separation tracking~\cite{svarny2019safe}, while exteroceptive RGB-D systems refine human position estimation near robotic manipulators~\cite{tashtoush2021human}. Skeletal tracking methods have also been employed to calculate minimum safe distances, reducing dependency on wearable safety systems~\cite{secil2022minimum}.
    
    Safety in dynamic, multi-human environments requires adaptive strategies beyond static metric-based approaches. Traditional models often fail in settings such as warehouses, search and rescue operations, and service applications. Existing frameworks include the Danger Index (DI), which integrates distance and velocity for trajectory planning~\cite{kulic2007pre}, and the Kinetostatic Danger Field (KDF) for real-time risk assessment~\cite{lacevic2013safety}.
    Palmieri et al. \cite{palmieri2024control} presented a human safety field (HSF) control architecture to improve safety in shared workspaces by adjusting manipulator trajectories. 
    Although these models improve regulatory compliance, they often neglect psychological and contextual factors, which can lead to discomfort, reduced trust, and limited adoption. Integrating human perception into safety assessments is essential for designing robots that are not only physically safe, but also intuitively comfortable for users.

\subsection{Perceived Physical Safety}

Perceived Physical Safety (PPS) refers to an individual's subjective assessment of physiological safety when interacting with a robot, influenced by trust, comfort, prior experience, and emotional state. Unlike absolute physical safety, PPS varies across contexts; individuals interacting with delivery robots may prefer greater distances, while those engaging with caregiver robots may tolerate closer proximity~\cite{rossi2017user}. Motion dynamics plays a crucial role, as unpredictable or abrupt movements can induce discomfort even when robots maintain objectively safe distances~\cite{nertinger2023influence}. Factors such as individual variability, environmental conditions, and motion predictability further shape PPS, with studies categorizing these influences into distinct models~\cite{akalin2023taxonomy}. Human predictability and reaction times also affect safety perception, necessitating robot control algorithms that align with behavioral responses~\cite{lasota2017survey}. 

Various approaches have been used to measure PPS, encompassing both subjective and objective methodologies. Subjective evaluations often employ standardized questionnaires, such as the GODSPEED (GS) \cite{bartneck2009measurement} and Robotic Social Attributes Scale (RoSAS) \cite{Carpinella} questionnaire series, which assesses dimensions like safety, comfort, trust, and perceived intelligence of robots. Other methods include interviews, open-ended surveys, and Likert-scale ratings that provide insights into user perceptions of safety and comfort in HRI scenarios. These approaches allow researchers to capture nuanced psychological states, typically in a post-hoc manner, that are difficult to quantify through objective measures alone. In addition, physiological measurements such as heart rate variability and skin conductance have been employed to provide objective correlates of perceived safety~\cite{akalin2023taxonomy}. 

Environmental and contextual factors have also been widely studied, with spatial configurations and robot behavior identified as key determinants of perceived safety~\cite{rosenfeld2019proactive}. Adaptive safety mechanisms integrating physical safety thresholds with real-time user feedback have been proposed to dynamically adjust safety parameters~\cite{shi2021augmented}, particularly in high-risk industrial settings. The need to combine subjective feedback with objective safety evaluations remains critical~\cite{akalin2022you}. Understanding the gap between absolute and perceived safety is essential for optimizing safety frameworks, as seen in pedestrian interactions with autonomous vehicles~\cite{seo2018perceived}. Situational awareness has also been identified as a fundamental factor influencing PPS~\cite{endsley1995toward}. Addressing these challenges requires an integrated approach considering both objective measures and human perception to improve real-world safety in HRI.

\subsection{Generalized Safety Index}
We leverage a real-time RGB-D-based system developed to measure safety during interactive tasks \cite{pandey2025freshrgsigeneralizedsafetymodel}, integrating a proxemics-guided \textbf{generalized safety index} (GSI). GSI quantifies perceived physical safety using relative distance and velocity between the human and the mobile robot to produce a real-time safety score. It accounts for the stopping distance in different proximity zones—intimate $(0-.46 m)$, personal $(.46-1.2 m)$, social $(1.2-3.7 m)$, and public $(>3.7 m)$ ~\cite{hall1966hidden,mumm2011human}, enabling proxemics-informed safety assessments. Let a mobile robot $r$ have pose $p_r = \langle \bm{x}_r, \theta_r \rangle$, where $\bm{x}_r = (x_r, y_r, z_r)$ is its position and $\theta_r$ its orientation. The robot operates alongside $N_h$ detected humans $\{ h_i \mid i = 1 \ldots N_h \}$, each at position $\bm{x}_{h_i} = (x_i, y_i, z_i)$ in a common reference frame. The Euclidean distance from the robot to human $h_i$ is defined as $d_{h_i,r} = \| \bm{x}_{h_i} - \bm{x}_r \|_2$, and the relative velocity is $v_{h_i,r} = -\dot{d}_{h_i,r}$, which is positive when the human approaches the robot. The relative bearing is given by $\theta_{h_i,r} = \measuredangle(\bm{x}_{h_i} - \bm{x}_r) - \theta_r$, representing the angle between the human’s position vector and the robot’s orientation. The GSI for a human $h_i$ is calculated as:
\begin{equation}
\begin{aligned}
    & {GSI}_{h_i}(d_{h_i,r},v_{h_i,r}; \rho) = \\ 
    & \text{clip}\Bigg( 
    \Bigg[ \frac{d_{h_i,r} - \left( \text{s}(v_{h_i,r}) \frac{v_{h_i,r}^2}{2 A_{\mathrm{max}}} + D_{\mathrm{min}} \right)}
    {D_{\mathrm{max}} - D_{\mathrm{min}}} \Bigg]^{\rho},\ 0,\ 1 \Bigg)
    \label{eqn:GSI_hat}
\end{aligned}
\end{equation}
where $A_{max}$ is the robot's maximum (de-)acceleration, $D_{max}$ denotes a safe distance (e.g., proxemics-guided $3.7m$ for public space), $D_{min}$ represents the minimum allowable distance (e.g., $.46m$ for intimate space), and $D_{max} > D_{min}$. The term $\frac{v_{h,r}^2}{2 A_{max}}$ represents the stopping distance based on the robot's current relative speed $v_{h,r}$. The sign function $\text{s}(v_{h,r})$ determines whether the human is approaching or moving away. The \textit{clip} ensures the resulting GSI values are bounded within $[0,1]$, clipping any values outside this range. A value between 0 and 1 measures the safety level -- closer to 0 indicates less safety and higher risk to the human at that point in time, whereas closer to 1 suggests that the human is likely to be safe at that time.

\begin{figure}[t] 
            \centering
            \includesvg[width=0.7\linewidth]{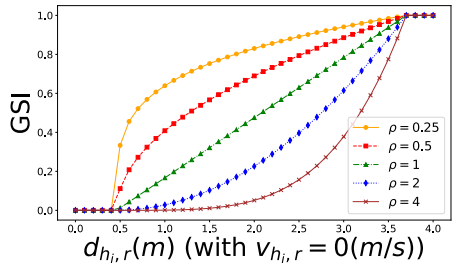}
            \vspace{-2mm}
            \caption{\small GSI can be fitted to various applications, robot platform properties, and subjective safety perceptions of humans through parameter $\rho > 0$. For instance, $\rho=1$ is set for assessing safety, $\rho>1$ for more cautious robot control, and $\rho<1$ for a closer interaction with humans who are already comfortable.}
            \label{fig:variation_graph}
            \vspace{-4mm}
        \end{figure}

The hyperparameter $\rho$ refines the perception of safety by modulating how humans psychologically assess risk and comfort beyond objective safety metrics. It can help contextualize GSI across different applications by incorporating human expectations, cognitive comfort, and environmental context. Lower values of $\rho$ indicate greater tolerance for close interactions, making them ideal for delivery and routine assistance, whereas higher values enforce stricter safety behavior, crucial for sensitive applications like healthcare. This adaptability allows robots to operate efficiently while maintaining user trust and comfort. Figure~\ref{fig:variation_graph} illustrates the impact of $\rho$ on GSI, showing how safety perception varies with decreasing distance at constant velocity.
By analyzing the hyperparameter $\rho$, we aim to understand the gap between APS and PPS, allowing inference on safety models to adapt to human-specific and environmental variations.

\section{Experiments}

\begin{figure}[t]
    \centering
    \includegraphics[width=0.95\linewidth]{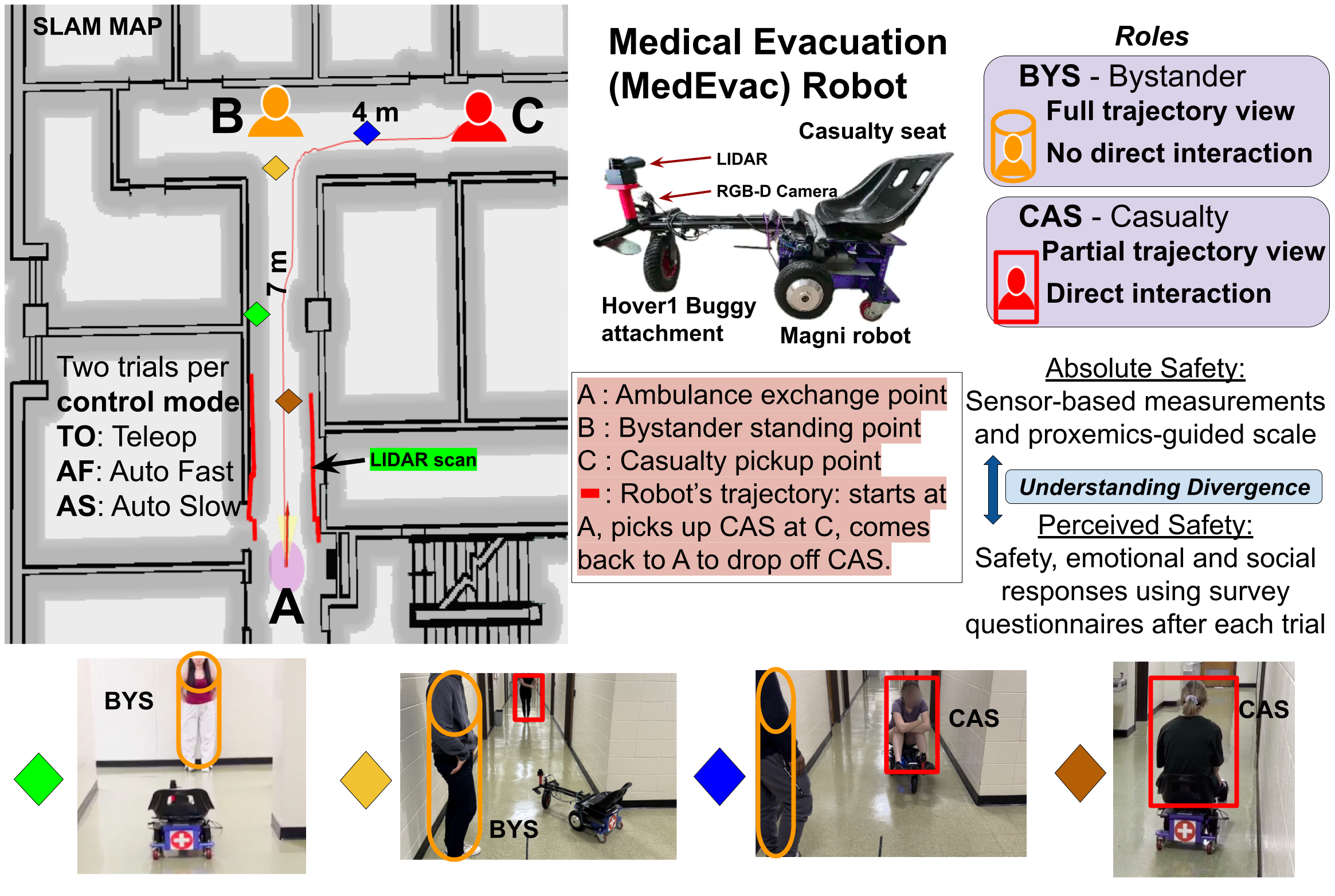}
    \caption{Simulated MEDEVAC scenario where BYS observe and CAS experience transport. The robot evacuates CAS from a pickup point to a drop-off location. Participants complete questionnaires after each trial to assess perceived safety, emotional states, and social compatibility.}
    \label{fig:overview}
    \vspace{-6mm}
\end{figure}

We employ a 2×3 mixed factorial design with one between-subject and one within-subject factor to assess human perceptions of an autonomous MEDEVAC robot~\cite{jordan2024analyzinghumanperceptionsmedevac}. The between-subject factor was the participant's role: bystanders (BYS) passively observe the robot while positioned between the casualty pickup and ambulance exchange point, simulating real-world non-interactive scenarios \cite{tsui}; casualties (CAS) were physically transported by the robot, representing direct human-robot interaction. The within-subject factor was the robot’s operating mode. Each participant experienced three modes: Autonomous-Slow (AS) at $.3m/s$, Autonomous-Fast (AF) at $.75m/s$, and Teleoperation (TO), where the robot was remotely controlled using a fixed speed at $.5m/s$ and by remotely viewing using the camera and lidar, without the participant’s awareness. This mode reflects the way most MEDEVAC robots are currently operated. All participants completed two trials per mode, totaling six trials per subject. The speeds for AS, AF, and TO were selected based on the robot’s capabilities and to reflect MEDEVAC constraints—simulating cautious, urgent, and human-controlled operation modes, respectively—and were validated through pilot testing for safety and stability. Figure~\ref{fig:overview} illustrates the MEDEVAC scenario including the robot’s route and the differing perspectives of BYS and CAS participants.

The physical setup involved a Ubiquity Magni robot equipped with a Hokuyo 2D LiDAR, Intel RealSense RGB-D camera, and a Hover-1 Beast platform for CAS transport. The autonomy stack, running ROS Noetic, utilized \texttt{move\_base} with A* for global planning and TEB for local optimization, while LIDAR-based SLAM ensured reliable navigation. 
We apply the FRESHR framework \cite{pandey2025freshrgsigeneralizedsafetymodel} using the RGB-D camera data to record the real-time distance and velocity of the detected human in the robot's travel path (i.e., BYS). Accordingly, the GSI values can be calculated for every trial when the $\rho$ is set.

\paragraph*{Participants} A total of 61 participants (77\% female, mean age = 20 years) from the University of Georgia, mainly from Psychology and Computer Science, were recruited. After the briefing, participants were assigned roles (30 BYS and 31 CAS) and completed six trials (two trials under each robot operating mode) in a randomized sequence to mitigate learning and ordering effects. The slight imbalance (31 CAS vs 30 BYS) resulted from participant unavailability; in such cases, a study assistant completed the dyad, but their data was excluded from analysis. The robot navigated an 11 m route between an “ambulance exchange” location and a casualty pickup point. BYS participants observed the robot without being transported, while CAS participants experienced transport. All trials were conducted consecutively, followed by a debriefing to gather feedback and ensure participant well-being.

\subsection{Measures}
\label{measures}
 Following each trial, participants completed questionnaires\footnote{Available at \url{https://tinyurl.com/mpdxv9rf}} assessing \textbf{emotional states (Q1), perceived safety (Q2-Q3), and social comfort (Q4)} regarding their recent experience with the MEDEVAC robot’s motion and proximity while it approaches them in their roles. 
 The questionnaire was trial-specific and thus completed by each subject six times (two trials per operating mode). 

\begin{itemize}
    \item Q1 \textcolor{black}{(Emotional State)}: Five 5-point Likert-scale items assess emotional states such as relaxation, calmness, quiescence, comfort, and predictability \cite{bartneck2009measurement}.
    \item Q2 \textcolor{black}{(General Perceived Safety)}: A single 7-point Likert item measures overall perceived safety from 'Unsafe' (1) to 'Overly Safe' (7), following \cite{van2023increasing}.
    \item Q3 \textcolor{black}{(Perceived Safety by Event)}: Five 5-point Likert items evaluate perceived safety across trial segments to capture temporal variation.
    \item Q4 \textcolor{black}{(Social Dimensions)}: Sixteen 5-point Likert items from the HMTI~\cite{grant_goodie_doshi_2024} measure social compatibility, including task completion, trust, and consistency .
\end{itemize}

\subsection{Estimating the safety hyperparameter}
\label{sec:MLE}
To quantitatively align sensor-based safety measures with perceived safety, we estimate the hyperparameter $\rho$ of the GSI model in \eqref{eqn:GSI_hat} through a maximum likelihood estimation (MLE) approach. Assuming that the distribution of error in reported perceived safety $(\widetilde{GSI}_i)$ is normal, the log-likelihood function simplifies to:
\begin{equation}
l(\rho) = -\frac{1}{2\sigma^2} \sum_{i=1}^n (\widetilde{GSI}_i - GSI_i(\rho))^2.
\label{eq:2}
\end{equation}
We apply the Broyden-Fletcher-Goldfarb-Shanno (BFGS)-based gradient search optimizer to obtain the optimum $\rho$, refining its estimation to align the GSI with participant-reported safety scores. We assume that the residual error between the reported safety and model-generated GSI represents white noise in human responses and follows a normal distribution. This assumption enables the application of the log-likelihood formulation in Eq.~\ref{eq:2} and provides a tractable method for estimating $\rho$ through MLE. The gradient descent update is performed as:
\begin{equation}
\rho \gets \rho + \eta \cdot \frac{\partial l(\rho)}{\partial \rho},
\end{equation}
where the learning rate ($\eta$ = 0.01) was empirically determined during initial tests for stable convergence using the BFGS optimizer. Through this approach, we derive individualized $\rho$ values, allowing us to tailor safety models that reflect diverse human perceptions in different interaction contexts. 
This data-driven approach enabled the personalization of the GSI, integrating individual safety perceptions into the model.

In particular, the estimation of $\rho$ was based on specific question items within Q3, tailored to each role. CAS responded to ``During the approach when the robot started from ‘B’ and moved towards the casualty at ‘C’,'' while BYS responded to ``During the robot’s motion from point ‘C’ to bystander location ‘B’.'' These questions are chosen because the GSI model applies only when the subject remains within the robot’s viewable range throughout the interaction. These questions pertain to regions of continuous human visibility, providing a reliable basis for $\rho$ estimation. Note that, by definition, we do not expect $\rho$ to be influenced by the operating modes (AF, AS, and TO). Thus, a single $\rho$ was learned from the data in all three modes. GSI is fitted in post hoc analysis only and did not influence the robot's behavior in real time. During the experiment, $\rho$ was fixed at 1 for all participants; personalized values were estimated offline after the study using Q3 responses and sensor data.

\subsection{Experiment Hypotheses}
\label{sec:hypotheses}

To better understand a human-centered, context-sensitive safety model, we formulated a set of hypotheses that primarily explored the effect of role on the robot's fitted safety index. These hypotheses capture key psychological, behavioral, and {role-based dynamics} of HRI in a MEDEVAC scenario. Each hypothesis provides a basis for predicting how different participants might respond, and together they inform the understanding of $\rho$ (representing individual differences in perceived safety).

\noindent \textbf{H1: Participants in both roles perceive the robot’s behavior as safe during MEDEVAC interactions.} \\
As long as the robot behaves in a {predictable} manner and maintains an {appropriate distance}, participants in both BYS and CAS roles will judge the interaction to be safe. Studies have shown that perceived safety is influenced by robot proximity, motion predictability, and trust in the robot’s behavior~\cite{bartneck2009measurement, lasota2017survey}.

\noindent \textbf{H2: Participants in both roles experience a positive emotional state while perceiving the interaction as safe.} \\
When participants report {positive emotional states} such as feeling calm, relaxed, or comfortable, they will also perceive the robot interaction as safe. Positive affective states are closely linked to higher safety perception~\cite{nertinger2023influence, akalin2023taxonomy}.

\noindent \textbf{H3: The MEDEVAC robot completing its task consistently makes participants in both roles trust the robot and perceive it as safe.} \\
Consistent task execution builds user trust and reduces perceived risk. Trust and predictability in the robot's performance are critical factors influencing perceived safety~\cite{hancock2011meta}.

\noindent \textbf{H4: Perception of safety is mediated by participant role. }\\
BYS are more likely to feel safer than CAS, as CAS participants anticipate close physical interaction while awaiting the robot for transportation, potentially increasing their sense of vulnerability. This is expected due to differing experiences of proximity, smoothness, and control~\cite{seo2018perceived}.

\noindent \textbf{H5: Repeated exposure to the robot enhances perceived safety over time across roles.} \\
The presence of multiple trials for each condition allows for this hypothesis, which investigates habituation effects (between the two trials in each condition): whether increased interaction frequency leads to greater comfort and reduced perceived risk \cite{schaefer2016meta}. While repeated exposure is expected to lead to higher safety and lower $\rho$, prior research also suggests that early impressions are durable \cite{willis2006first}, highlighting the importance of initial robot behavior.

\noindent \textbf{H6: The variability of perceived safety is captured using a few distinct clusters in both roles.} \\
In contrast to a high degree of individual variation in safety perception, we hypothesize that responses cluster into identifiable groups based on shared safety assessment and prior experiences with robots \cite{leichtmann2020how}. This enables the development of adaptive models that can personalize $\rho$ based on user {\em type}.

Collectively, these hypotheses offer a theoretical foundation for our safety evaluation. They help better understand the measurements and the effects we expect to see in the data. Together, the hypotheses span both objective factors (e.g., robot speed, task consistency) and subjective human factors (emotion, trust, role), ensuring that our analysis considers a comprehensive range of influences on perceived safety.

\section{Results}

First, we present the empirical findings of our hypothesis-driven study, focusing on how the estimated $\rho$ values capture variations in perceived safety across different human roles in the MEDEVAC scenario. 
The $\rho$ values are obtained for each participant via MLE described in Sec.~\ref{sec:MLE} by aligning the Q3 responses and the GSI with the recorded sensor data, leveraging the variations in human-robot relative distance and velocities of the three operating modes. 

\begin{table}[!ht]
    \centering
    \renewcommand{\arraystretch}{1.2}
    \setlength{\tabcolsep}{6pt}
    \begin{tabular}{|c|c|c|c|c|}
        \hline
        \textbf{Role} & \textbf{GSI (mean $\pm$ SE)} & \textbf{$\rho$ (mean $\pm$ SE)} & \textbf{Max $\rho$} \\
        \hline
        CAS & 0.85 $\pm$ 0.006 & 0.29 $\pm$ 0.05 & 1.20 \\
        BYS & 0.94 $\pm$ 0.003 & 0.97 $\pm$ 0.17 & 2.87 \\
        \hline
    \end{tabular}
    \caption{\small Summary statistics of learned GSI and estimated $\rho$ with Standard Error (SE) across CAS and BYS roles. The $\rho$ values reflect personalization in perceived safety: lower values indicate greater tolerance to close human-robot proximity.}
    \label{tab:rho_cas_bys_ststs}
    \vspace{-2mm}
\end{table}

Table~\ref{tab:rho_cas_bys_ststs} shows that participants in the CAS role had lower $\rho$ values than those in the BYS role, suggesting that individuals to be transported by the robot were more comfortable and felt safer likely because they anticipated close proximity. Both participant groups exhibited $\rho$ values less than one (see Fig.~\ref{fig:variation_graph}), indicating a higher perception of safety and general willingness to tolerate close proximity interactions. 

\begin{figure}[t]
    \centering
    \includesvg[width=0.8\linewidth]{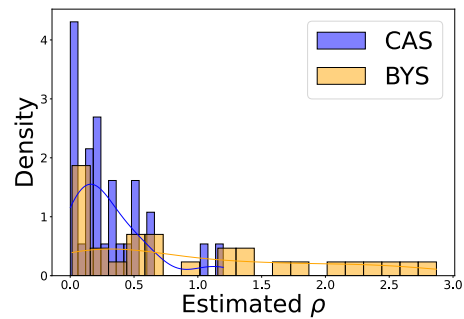}
    \caption{\small Histogram with kernel density estimate (KDE) overlay showing the distribution of estimated $\rho$ values for both BYS and CAS roles. Lower values suggest a higher perception of safety and acceptance of closer proximity.}
    \label{fig:KDE}
    \vspace{-4mm}
\end{figure}

Figure~\ref{fig:KDE} further illustrates this trend: both groups’ $\rho$ distributions are skewed toward lower values indicating bias toward higher perceived safety, although the BYS responses span a wider range, underscoring greater individual variability likely influenced by observational distance and prior experiences. These observations indicate a generally positive safety assessment and hint at role-specific differences in safety perception. Participant responses to questions related to emotional states, perceived safety, and task consistency are available on request for all trials in each control mode, separated by role. Notably, the responses are summarized in Table~\ref{tab:hypotheses_summary}. For instance, higher calmness and trust ratings during robot motions correspond to lower estimated $\rho$ trends, offering initial empirical support for our hypotheses.

\begin{table}[t]
\centering
\renewcommand{\arraystretch}{1.2}
\resizebox{\columnwidth}{!}{%
\begin{tabular}{|c|l|c|c|c|c|c|}
\hline
\textbf{Hyp} & \textbf{Question} & \multicolumn{2}{c|}{\textbf{Responses (Mean $\pm$ SD)}} & \multicolumn{3}{c|}{\textbf{Correlation ($r$) / Significance}} \\ \hline
& & \textbf{BYS} & \textbf{CAS} & \textbf{Type} & \textbf{BYS} & \textbf{CAS} \\ \hline
H1 & Q2 General PPS & 5.46 $\pm$ 1.34 & 4.23 $\pm$ 1.20 & Pearson & 0.446* & 0.51** \\ \hline
\multirow{5}{*}{H2} & Q1 Relaxed & 4.19 $\pm$ 1.11 & 3.85 $\pm$ 1.01 & Pearson & 0.38* & 0.69*** \\
   & Q1 Calm & 4.34 $\pm$ 0.88 & 4.07 $\pm$ 0.93 & Pearson & 0.61*** & 0.39* \\
   & Q1 Quiscent & 3.49 $\pm$ 1.45 & 3.53 $\pm$ 1.08 & Pearson & 0.33 & 0.44** \\
   & Q1 Comfortable & 4.42 $\pm$ 0.98 & 3.91 $\pm$ 0.95 & Pearson & 0.42* & 0.37* \\
   & Q1 Predictable & 3.74 $\pm$ 1.33 & 3.44 $\pm$ 1.07 & Pearson & 0.49** & 0.54** \\ \hline
\multirow{3}{*}{H3} & Q4 Completion & 4.67 $\pm$ 0.61 & 4.20 $\pm$ 1.13 & Pearson & 0.53** & 0.37* \\
   & Q4 Trust & 4.19 $\pm$ 0.96 & 3.20 $\pm$ 1.20 & Pearson & 0.29 & 0.49** \\
   & Q4 Consistency & 3.99 $\pm$ 1.00 & 3.18 $\pm$ 1.15 & Pearson & 0.37* & 0.38* \\ \hline
H4 & Q3 & 4.29 $\pm$ 0.81 & 3.85 $\pm$ 1.02 & Mann Whitney & \multicolumn{2}{c|}{0.36***} \\ \hline
\multirow{2}{*}{H5 Trialwise} & Q3 Trial 1 & 4.29 $\pm$ 0.72 & 3.58 $\pm$ 0.99 & \multirow{2}{*}{Wilcoxon} & \multirow{2}{*}{0.69} & \multirow{2}{*}{0.25***} \\
             & Q3 Trial 2 & 4.30 $\pm$ 0.90 & 4.11 $\pm$ 0.98 &  & & \\ \hline
\end{tabular}%
}
\caption{\small Summary of the participant response data. Scores are reported as mean $\pm$ SD. Higher scores on the Likert scale indicate stronger (positive) alignment with the corresponding metric. $*$ indicates significance level $(^*p < 0.05, ^{**}p < 0.01, ^{***}p < 0.001)$. Effect sizes are reported for all correlation and hypothesis tests. Per Cohen's guidelines~\cite{cohen2013statistical}, $r = 0.1$ is considered a small effect, $r = 0.3$ a medium effect, and $r \geq 0.5$ a large effect.}

\label{tab:hypotheses_summary}
\vspace{-3mm}
\end{table}

To formally evaluate \textbf{H1} — that participants in both roles would perceive the robot’s trajectories as safe — we examined the correlation between participants’ subjective safety ratings and the GSI computed using the estimated $\rho$. Participants rated each trajectory on a 7-point Likert scale from –3 (unsafe) to +3 (overly safe), which we normalized to a 5-point scale for consistency. The analysis revealed (see Table~\ref{tab:hypotheses_summary}) a statistically significant positive correlation between these safety ratings and the GSI in both roles (Pearson’s $r$ significant with $p<0.01$). This is indicative of the goodness of the fit of the empirically-informed GSIs and the validity of the learned $\rho$ values. These tend to be low, 
confirming that participants generally found the robot’s motion to be safe. This relationship was somewhat stronger for BYS participants than for CAS participants. 
despite BYS observers displaying slightly more variability. Collectively, this finding validates H1 and reinforces that our calibrated parameter $\rho$ effectively captures subjective safety perceptions during the interaction.

Beyond overall safety ratings, we also hypothesize that participants’ affective states would shape their safety perceptions. To evaluate \textbf{H2} — that participants in both roles would feel a positive emotional state (calm, relaxed, and comfortable) when the interaction is perceived as safe — we examined the relationship between reported emotional states and the corresponding GSI values with estimated $\rho$. Emotional responses were measured using semantic differential scales (e.g., Anxious–Relaxed, Agitated–Calm, Uncomfortable–Comfortable, Unpredictable–Predictable, Surprised–Quiescent) derived from the GODSPEED questionnaire~\cite{bartneck2009measurement}  and related literature. Table~\ref{tab:hypotheses_summary} shows significant correlations between emotional states and perceived safety, particularly for the CAS group.

The strongest correlation was observed on the Anxious–Relaxed scale, where feeling relaxed strongly correlated with a greater sense of safety (Pearson’s $r=0.699$, $p<0.001$ for CAS). Significant positive correlations were also found for the CAS group on the Agitated–Calm, Uncomfortable–Comfortable, and Unpredictable–Predictable scales (all $p<0.01$), indicating that as the robot was perceived as more calming and its behavior more predictable, participants felt safer. The BYS group showed a similar trend of positive correlations between emotional comfort and safety, albeit generally weaker.

In summary, the results support \textbf{H2}: Participants’ positive emotional states (greater calm, comfort, and predictability) were strongly associated with higher perceived safety (lower $\rho$). This finding reinforces the importance of incorporating human affect into safety models, as those who felt at ease and in control aligned with perceiving the interaction as safer.

To evaluate \textbf{H3} — that participants would perceive the robot as safe if the robot consistently and successfully completed its task — we analyzed how participants’ trust and performance ratings related to their safety impressions. Participants rated their agreement (on a 5-point Likert scale) with several statements on the robot's reliability and task performance. Pearson correlation analysis was then conducted between these ratings and the GSI using estimated $\rho$. 

We found significant positive correlations between perceived task performance and GSI for both groups (see Table~\ref{tab:hypotheses_summary}). Specifically, those who agreed that \textit{the robot consistently performs well} or \textit{completed its tasks successfully} exhibited elevated GSI scores. The strongest correlation was with the statement “I trusted the machine to perform well as the task went on” for CAS participants ($r=0.49$, $p<0.01$). For BYS participants, the highest correlation was observed with the statement “This machine completed its tasks successfully” ($r=0.53$, $p<0.01$). These findings confirm \textbf{H3}.

To evaluate \textbf{H4} — that perceived safety is mediated by participant role — we compared GSI using learned values of $\rho$ between participants in CAS and BYS. As shown in Table~\ref{tab:rho_cas_bys_ststs}, CAS participants generally exhibited lower $\rho$ values, indicating greater tolerance for close proximity, while BYS participants reported higher values. Moreover, GSI using the estimated $\rho$ also shows this difference as we may expect. A Mann-Whitney U test revealed a significant effect of role on learned GSI values ($r = 0.36$, $p < 0.001$), confirming that safety perceptions differ systematically based on the participant's position and level of engagement during the robot interaction.

\begin{figure*}[t]
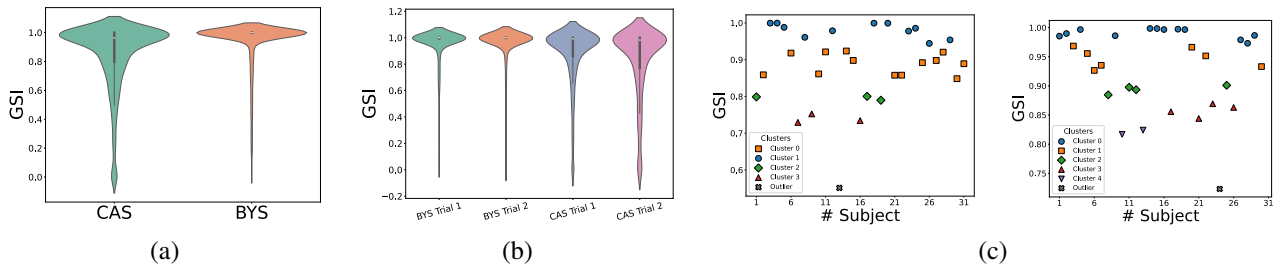

    \centering
    \begin{tabular}{cccc}
        \includesvg[width=0.24\linewidth]{FIGURES/GSI_violin_plot.svg} & 
        \includesvg[width=0.24\linewidth]{FIGURES/GSI_violin_plot_trial.svg} &
        \includesvg[width=0.2\linewidth]{FIGURES/CAS_CAS_MeanShift_clustering_perception_Safety_GSI.svg} &
        \includesvg[width=0.2\linewidth]{FIGURES/BYS_CAS_MeanShift_clustering_perception_Safety_GSI.svg} \\
        (a) & (b) & \multicolumn{2}{c}{(c)} \\
    \end{tabular}
    \vspace{-2mm}
    \caption{\small 
    (a) Violin plot comparing GSI with learned $\rho$ between BYS and CAS roles, showing higher variability and median for CAS. 
    (b) Trial-wise distribution of GSI with learned $\rho$ in both roles, showing stable perception across time for BYS but not for CAS. 
    (c) Mean-Shift clustering of GSI values using learned $\rho$ values for CAS (left) and BYS (right), revealing greater perceptual variability among BYS.
    }
    \label{fig:combined_results}
    \vspace{-4mm}
\end{figure*}

Figure~\ref{fig:combined_results}(a) illustrates the distribution of GSI scores across roles. CAS participants exhibit greater variability in GSI with estimated $\rho$ compared to BYS participants, whose responses are tightly clustered near the upper bound. This divergence indicates that while BYS tend to maintain consistently high safety margins—likely due to their limited engagement and passive observation—CAS participants show a broader range, possibly reflecting individualized adjustments stemming from direct physical interaction with the robot. These findings align with and support \textbf{H4}, confirming a significant role-based difference in safety perception.

To evaluate \textbf{H5} — that repeated exposure to the robot enhances perceived safety — we compared GSI across trials for both roles separately (Fig.~\ref{fig:combined_results}(b)). BYS participants showed no significant change (Trial~1: $0.94 \pm 0.004$, Trial~2: $0.94 \pm 0.004$, Wilcoxon $r = 0.69$, $p = 0.30$), indicating stable safety perception. However, CAS participants experienced a significant decrease (Trial~1: $0.88 \pm 0.008$, Trial~2: $0.83 \pm 0.011$, Wilcoxon $r = 0.25$, $p < 0.001$), suggesting that repeated exposure led to reduced perceived safety. These results are consistent with the descriptive statistics in Table~\ref{tab:descriptive_stats}.

\begin{table}[ht]
    \centering
    \renewcommand{\arraystretch}{1.2}
    \setlength{\tabcolsep}{4pt}
    \begin{tabular}{|c|c|c|c|}
        \hline
        \textbf{CAS} & \textbf{GSI (mean $\pm$ SE)}  & \textbf{BYS} & \textbf{GSI (mean $\pm$ SE)}  \\ 
        \hline
        Trial 1 & 0.88 $\pm$ 0.008 & Trial 1 & 0.94 $\pm$ 0.004\\
        Trial 2 & 0.83 $\pm$ 0.011 & Trial 2 & 0.94 $\pm$ 0.004\\ 
        \hline
    \end{tabular}
    \caption{\small Descriptive statistics for GSI using estimated $\rho$ values across trials. The consistency in mean GSI values indicates that perceived safety remained stable despite repeated exposure.}
    \label{tab:descriptive_stats}
\end{table}

Beyond group differences by role, we also explored whether individual variations in perceived safety would form distinct clusters of respondents. To investigate \textbf{H6}, we applied the mean-shift clustering algorithm to the GSI values with  estimated $\rho$ for each role (see Fig.~\ref{fig:combined_results}(c)). This approach revealed differences between the two groups. CAS participants formed broader and more dispersed clusters, suggesting lower variability. In contrast, BYS participants formed a greater number of tighter clusters, indicating a more diverse set of safety perceptions. Table~\ref{tab:clustering_results} shows higher perceptual variability in BYS participants, compared to CAS. These findings demonstrate that perceived safety is not uniformly distributed but instead forms distinct perception groups. The higher number and cohesion of BYS clusters may result, in part, from varying interpretative perspectives. In contrast, CAS participants, having more proximal interaction with the robot, displayed a more consistent perception of safety.

\begin{table}[ht!]
    \centering
    \renewcommand{\arraystretch}{1.2}
    \setlength{\tabcolsep}{4pt}
    \resizebox{\columnwidth}{!}{%
    \begin{tabular}{|c|c|c|c|c|c|c|c|}
        \hline
        \multirow{2}{*}{\textbf{Role}} & \textbf{} & \multicolumn{6}{c|}{\textbf{Cluster}} \\
        \cline{3-8}
        & & \textbf{0} & \textbf{1} & \textbf{2} & \textbf{3} & \textbf{4} & \textbf{Outliers} \\
        \hline
        \multirow{2}{*}{CAS} & \#Datapoints & 13 & 11 & 3 & 3 & $-$ & 1  \\
        & GSI & $0.89 \pm 0.03$ & $0.98 \pm 0.02$ & $0.80 \pm 0.00$ & $0.74 \pm 0.01$ & $-$ & $0.55 \pm 0.00$ \\ 
        \hline
        \multirow{2}{*}{BYS} & \#Datapoints & 12 & 7 & 4 & 4 & 2 & 1 \\
        & GSI & $0.99 \pm 0.01$ & $0.95 \pm 0.02$ & $0.89 \pm 0.01$ & $0.86 \pm 0.01$ & $0.82 \pm 0.00$ & $0.72 \pm 0.00$ \\
        \hline
    \end{tabular}
    }
    \caption{\small Cluster distribution of GSI using estimated $\rho$ values for CAS and BYS participants. The number of participants per cluster is shown under each cluster number. BYS exhibited a higher silhouette score, reflecting greater variability in perceived safety.}
    \label{tab:clustering_results}
\end{table}

While these findings provide a detailed understanding of safety perception in multi-human HRI, a deeper examination of the underlying cognitive and behavioral mechanisms is necessary. The following discussion section critically evaluates these insights, contextualizing them within existing literature and exploring their implications for future advancements in human-aware robotic systems.

\section{Discussion}

\textbf{H1: Confirmed} Analysis of the GSI values using estimated $\rho$ demonstrates strong correlation with perceived safety. Specifically, BYS participants exhibited greater correlation to safety concerns compared to CAS participants. This observation supports prior research indicating that safety perception is influenced by both objective factors (e.g., distance, velocity) and subjective factors (e.g., familiarity, predictability)~\cite{nertinger2023influence}. The higher GSI values for CAS participants indicate a higher tolerance for close proximity, which aligns with the \textit{trust calibration} theory~\cite{rossi2017user}. The CAS group, having direct interaction with the robot, developed situational trust, while BYS participants, relying on visual cues, demonstrated more cautious tendencies. This finding highlights the need for adaptive safety mechanisms that consider role-specific factors and provide context-related safety.

\textbf{H2: Confirmed} Observations indicate that GSI values are significantly correlated with participants' emotional states, such as calmness, comfort, and predictability. CAS participants demonstrated greater correlation to motion predictability than BYS participants, consistent with findings that humans prefer smooth, controlled robotic movements~\cite{leichtmann2020how}. Additionally, this supports \textit{cognitive appraisal theory}~\cite{lazarus1984stress}, where initial emotional responses critically influence perceived safety. When robots maintain predictable trajectories, GSI values indicate higher perceived safety, particularly for CAS participants. Strategies that improve motion predictability and trajectory clarity can enhance perceived safety across roles.

\textbf{H3: Confirmed} Correlation analyses between GSI values and perceived task consistency reveal that participants' trust in the robot's performance influences their safety perception. Both groups exhibited stronger correlation to reliability, suggesting that irrespective of role, consistently completing its task successfully plays a significant role in evaluating safety. This finding aligns with \textit{predictive processing models}~\cite{clark2013whatever}, where consistent and successful interactions reinforce trust, resulting in higher GSI values. Conversely, inconsistencies or failures degrade perceived safety, highlighting the importance of behavioral consistency in improving GSI scores.

\textbf{H4: Confirmed} Differences in perceived safety between CAS and BYS roles underscore the influence of physical interaction. CAS participants, who are awaiting direct contact with the robot, likely faced heightened sensitivity and trust calibration, consistent with findings that physical embodiment impacts perceived safety \cite{rubagotti2022perceived}. In contrast, BYS participants showed more stable and conservative perceptions, aligning with results suggesting passive observation leads to more consistent but less emotionally salient responses \cite{hancock2011meta}.

\textbf{H5: Not Confirmed} While short-term repeated exposure to the robot did not appreciably shift safety perceptions in BYS participants—mirroring findings that passive observation offers limited sensory feedback for internal safety model updates \cite{dragan2015effects}—CAS participants exhibited a statistically significant decline in perceived safety across trials. This supports existing literature suggesting that direct physical interaction, especially in close proximity, can increase user sensitivity or discomfort when inconsistencies arise \cite{choi2017movement}. These findings caution against assuming that repeated exposure will always enhance trust or comfort in embodied robotic interactions.

\textbf{H6: Confirmed} Clustering analysis of GSI values reveals that perceived safety is not continuous but rather organized into discrete psychological profiles. The BYS group exhibited greater perceptual variability, forming several distinct clusters, whereas the CAS group demonstrated a more cohesive perception. This observation supports \textit{cognitive categorization}~\cite{akalin2023taxonomy}, where human responses to robotic interactions fall into identifiable groups. Modeling safety perception through clustering provides a structured framework to adjust robot behavior based on categorized safety profiles dynamically.


\section{Conclusion}

This paper presented a data-driven analysis of perceived physical safety in human-robot interaction, grounded in subjective human evaluations and objective proximity data. We introduced a parameterized formulation of the Generalized Safety Index (GSI), incorporating a personalization factor, $\rho$, to better align safety metrics with individual perceptions during robot approach behaviors. Experiment results in a MEDEVAC context revealed the significance of motion predictability and role-based exposure in shaping perceived safety, with bystanders reporting higher safety than those awaiting robotic evacuation. These findings highlight the value of tailoring physical safety assessments to human perspectives and offer empirical insights into perception-driven safety modeling. Our work lays the groundwork for future efforts to integrate human feedback more deeply into robotic behavior evaluation in complex social environments.

\section*{Acknowledgements}
This research was sponsored by the Army DEVCOM Analysis Center under Cooperative Agreement W911NF-22-2-0001. The views expressed are those of the authors and do not reflect official policies of the DEVCOM Analysis Center or the U.S. Government. 
The U.S. Government may reproduce and distribute reprints for official purposes, regardless of copyright notice.
We also thank Kenneth Bogert for his assistance with robot setup and coding during the initial stages of the experiment, and Tyson Jordan for his assistance with the human subject experiments.


\bibliographystyle{IEEEtran}
\bibliography{main}

\end{document}